%% file: acl_latex.tex
\pdfoutput=1

\documentclass[11pt]{article}

\usepackage[final]{acl}

\usepackage{times}
\usepackage{latexsym}

\usepackage[T1]{fontenc}

\usepackage[utf8]{inputenc}

\usepackage{microtype}

\usepackage{inconsolata}

\usepackage{graphicx}
\usepackage{inconsolata}
\usepackage{xcolor}
\usepackage{enumitem}
\usepackage{booktabs}
\usepackage{array}
\usepackage{hyperref}
\usepackage{url}
\usepackage{multirow}
\usepackage{colortbl}
\usepackage{booktabs}
\usepackage{amssymb}
\usepackage{amsmath}
\usepackage{amsthm}
\usepackage{graphicx}
\usepackage{makecell}
\usepackage{threeparttable}
\usepackage{CJKutf8}
\usepackage{lscape}
\usepackage{fancyhdr}
\usepackage[ruled,vlined]{algorithm2e}

\usepackage{threeparttable}
\usepackage{algorithmic}
\usepackage{wrapfig}
\usepackage{subfigure}

\title{Capability Salience Vector: Fine-grained Alignment of Loss and Capabilities for Downstream Task Scaling Law}

\author{
    {\normalsize
     \textbf{Qiming Ge}$^{\heartsuit\bigstar*}$, 
     \ \ Shuhao Xing$^{\heartsuit\bigstar}$\thanks{Equal contribution. This work was done when Qiming Ge was an intern at Shanghai AI Laboratory.},
     \ \ Songyang Gao$^{\heartsuit}$, 
     \ \ Yunhua Zhou$^{\heartsuit}$,
     }\\
    {\normalsize
     \textbf{Yicheng Zou}$^{\heartsuit}$\thanks{Corresponding Author}, 
    \ \ \textbf{Songyang Zhang}$^{\heartsuit}$,
    \ \ \textbf{Zhi Chen}$^{\heartsuit}$,
    \ \ \textbf{Hang Yan}$^{\heartsuit}$,
    \ \ \textbf{Qi Zhang}$^{\bigstar}$,
    }\\
    {\normalsize
     \textbf{Qipeng Guo}$^{\heartsuit}$, 
    \ \ \textbf{Kai Chen}$^{\heartsuit}$
    }\\
    {$^\heartsuit$Shanghai AI Laboratory} \\
    {$^\bigstar$College of Computer Science and Artificial Intelligence, Fudan University} \\
    \texttt{qmge22@m.fudan.edu.cn}, \texttt{zouyicheng@pjlab.org.cn}
}

\begin{document}
\maketitle
\begin{abstract}
Scaling law builds the relationship between training computation and validation loss, enabling researchers to effectively predict the loss trending of models across different levels of computation. However, a gap still remains between validation loss and the model's downstream capabilities, making it untrivial to apply scaling law to direct performance prediction for downstream tasks. The loss typically represents a cumulative penalty for predicted tokens, which are implicitly considered to have equal importance. Nevertheless, our studies have shown evidence that when considering different training data distributions, we cannot directly model the relationship between downstream capability and computation or token loss. To bridge the gap between validation loss and downstream task capabilities, in this work, we introduce \textbf{Capability Salience Vector}, which decomposes the overall loss and assigns different importance weights to tokens to assess a specific meta-capability, aligning the validation loss with downstream task performance in terms of the model’s capabilities. Experiments on various popular benchmarks demonstrate that our proposed Capability Salience Vector could significantly improve the predictability of language model performance on downstream tasks.
\end{abstract}

\input{section/1_introduction}

\input{section/2_relatedwork}
\input{section/3_method}

\input{section/4_experiment}

\input{section/5_conclusion}

\input{section/6_limitation}

\bibliography{custom}

\input{section/appendix}

\end{document}

%% file: section/1_introduction.tex
\section{Introduction}

Large language models have demonstrated impressive performance across a wide range of tasks, but this achievement comes with the trade-off of significant computational demands. To mitigate the computational burden during the training of large models, researchers have developed the scaling law \citep{bahri2024explaining, hoffmann2022training, kaplan2020scaling, muennighoff2024scaling}, a framework that predicts how variations in model size and data scale impact the validation loss of LLMs. This allows researchers to use smaller models to predict the validation loss of larger ones, reducing the cost of hyperparameter tuning and trial-and-error experimentation. However, in practical applications, a model's capabilities are typically assessed using a variety of downstream benchmarks, and validation loss does not always correlate directly with performance on these tasks. It is not uncommon to encounter instances where models with similar validation losses exhibit significantly different downstream task performance \cite{liu2023same}. Consequently, validation loss alone is not always a reliable indicator of a model’s true capabilities.

To address this issue, some researchers have chosen to bypass validation loss, focusing instead on directly establishing the scaling relationship between computational resources and task performance \citep{gadre2024language}. To minimize the effects of data distribution shifts, they adjusted the amount of computation and model sizes while keeping the pre-training dataset fixed across all experimental settings. However, we argue that the impact of data distribution shifts cannot be overlooked, as they may substantially influence model performance on downstream tasks. As demonstrated by our experiments (Section \ref{sec:c_to_p}), the same amount of computation can yield markedly different outcomes when models are trained on distinct data distributions.

Another line of research focuses on developing alternative metrics that better correlate with downstream task performance. For instance, Ruan et al. \shortcite{ruan2024observational} proposed the Observational Scaling Law (OSL), leveraging open-source benchmark results and applying principal component analysis (PCA) to explore various dimensions of model capabilities. While these approaches encompass a broad range of model families that are pre-trained on different data distributions, they depend heavily on the diversity and scope of the benchmarks used, which can incur substantial evaluation costs.

From the discussions above, we identify two important but underexplored questions that this work seeks to address: 
\begin{itemize}
    \item Is it reasonable to predict downstream capabilities directly based on computational cost?
    \item Why is it challenging to establish a clear relationship between validation loss and a model's capabilities?
\end{itemize} 
To address the first question, we explored the impact of different pre-training data distributions on the ability to model downstream tasks. Our experiments show that, under the same computational budget, different pre-training data distributions can lead to varying downstream capabilities. Therefore, it is not reasonable to predict downstream task performance directly based on computational cost.
To answer the second question, we revisit the concept of validation loss, a metric used to evaluate a model's cumulative penalty for next-token predictions. This metric inherently assumes that all tokens contribute equally to the model's training process. However, previous studies have demonstrated that tokens may differ in their learnability and difficulty \cite{lin2024rho, xia2022training}. Inspired of this, we introduce the concept of \textbf{Capability Salience Vector} (CSV), which quantifies the capabilities required for downstream tasks by assigning different importance weights to tokens. By computing a weighted average of the loss for these tokens, this metric can replace the raw validation loss and demonstrate a stronger correlation with downstream task performance. Furthermore, we propose an optimization algorithm to automatically derive the Capability Salience Vector for any given text. Experiments conducted on six popular benchmarks show that our approach consistently achieves a high correlation across different model series and significantly enhances the predictability of language model performance on downstream tasks.

Overall, our contributions are threefold:

\begin{itemize}
    \item  Our experiments indicate that downstream scaling laws cannot be directly modeled using computation or the average token loss on the validation set.

    \item We introduce the Capability Salience Vector and present an algorithm to automatically derive it. This allows us to quantify a model's capabilities by assigning different importance weights to the loss of specific tokens, thereby establishing a strong correlation between validation loss and downstream task performance.

    \item Through experiments on six benchmarks for different model capabilities, we show that our method consistently achieves high correlation across various model series and significantly improves the predictability of 
    model performance on these benchmarks. The predicted mean squared error was maintained within the range of 1e-3.

\end{itemize}

%% file: section/2_relatedwork.tex
\section{Related work}

\subsection{Scaling Law for Downstream Tasks}

Scaling law establishes the power-law relationships between computation $C$, model parameters $N$, the number of training tokens $D$, and cross-entropy loss on the validation set. However, there is still a gap between validation loss and downstream task performance. Recent studies have tried to investigate the scaling laws governing downstream task performance. Owen \shortcite{owen2024predictable} investigated the predictability of model scaling on BBH \cite{srivastava2022beyond} and MMLU \cite{hendrycks2020measuring} tasks, proposing that downstream task performance follows a sigmoidal function and can be predicted. Their findings also indicate that prediction errors decrease as more models are incorporated into the fitting process. Ruan et al. \shortcite{ruan2024observational} introduced an alternative observational approach, applying low-rank decomposition to publicly available model evaluation results to extract sub-dimensions representing the model capabilities. These sub-dimensions were then used to establish scaling laws that relate computational cost to downstream task performance. Gadre et al. \shortcite{gadre2024language} examined the scaling laws of overtraining models on specific pre-training data distributions. They identified a power-law relationship between validation loss and downstream task performance, demonstrating predictability after pre-training on a given data distribution.

However, the aforementioned studies primarily focused on the relationship between computational cost and downstream task performance, without addressing the fact that different pre-training data distributions can result in different downstream performances for the same computational cost. Isik et al. \shortcite{isik2024scaling} explored the impact of pre-training data distribution on downstream performance in translation tasks. Their findings revealed that different pre-training distributions significantly influence the scaling behavior of downstream tasks, indicating that data distribution should be considered when evaluating model scaling laws.

\subsection{Efficient Benchmark Prediction} 

As large language models continue to advance, numerous benchmarks have been developed to assess their diverse capabilities. To comprehensively evaluate the performance of a pre-trained model, it is necessary to measure its scores across various downstream tasks. However, conducting a full downstream evaluation is often time-consuming, prompting recent efforts to improve its efficiency. For example, Ye et al. \shortcite{ye2023predictable} and Perlitz et al. \shortcite{perlitz2023efficient} have explored reducing the number of task examples in benchmarks like BBH \cite{srivastava2022beyond} and HELM \cite{liang2022holistic} to streamline the evaluation process.

Recent studies, such as Polo et al. \shortcite{polo2024tinybenchmarks} and Kipnis et al. \shortcite{kipnis2024texttt}, have explored leveraging Item Response Theory \cite{lord2008statistical, baker2001basics} to extract model features from open-source evaluations and compress validation sets. This allows for the reconstruction of benchmark scores using these compressed subsets. Similarly, Pacchiardi et al. \shortcite{pacchiardi2024100} applied a feature extraction approach and trained an evaluator to directly predict performance on individual samples. Zhang et al. \shortcite{zhang2024collaborative} demonstrated that incorporating task performance across different model families, alongside additional model and task information, can enhance prediction accuracy. However, these methods still rely on accuracy metrics for part of the data, and it remains uncertain whether downstream task performance (e.g., accuracy) can be directly predicted based on downstream scaling laws.

%% file: section/3_method.tex
\section{Method}
In this section, we first explore whether downstream task performance can be directly predicted using computation or average token loss. Then, we introduce \textbf{Capability Salience Vector} (CSV), which automatically assign different importance weights to validation token loss. This aligns the loss of tokens with the specific capabilities required for downstream tasks.

\subsection{Impact of Pre-training Data Distribution on Modeling Downstream Scaling Law}\label{sec:c_to_p}
\begin{figure}[ht]
\includegraphics[width=\linewidth]{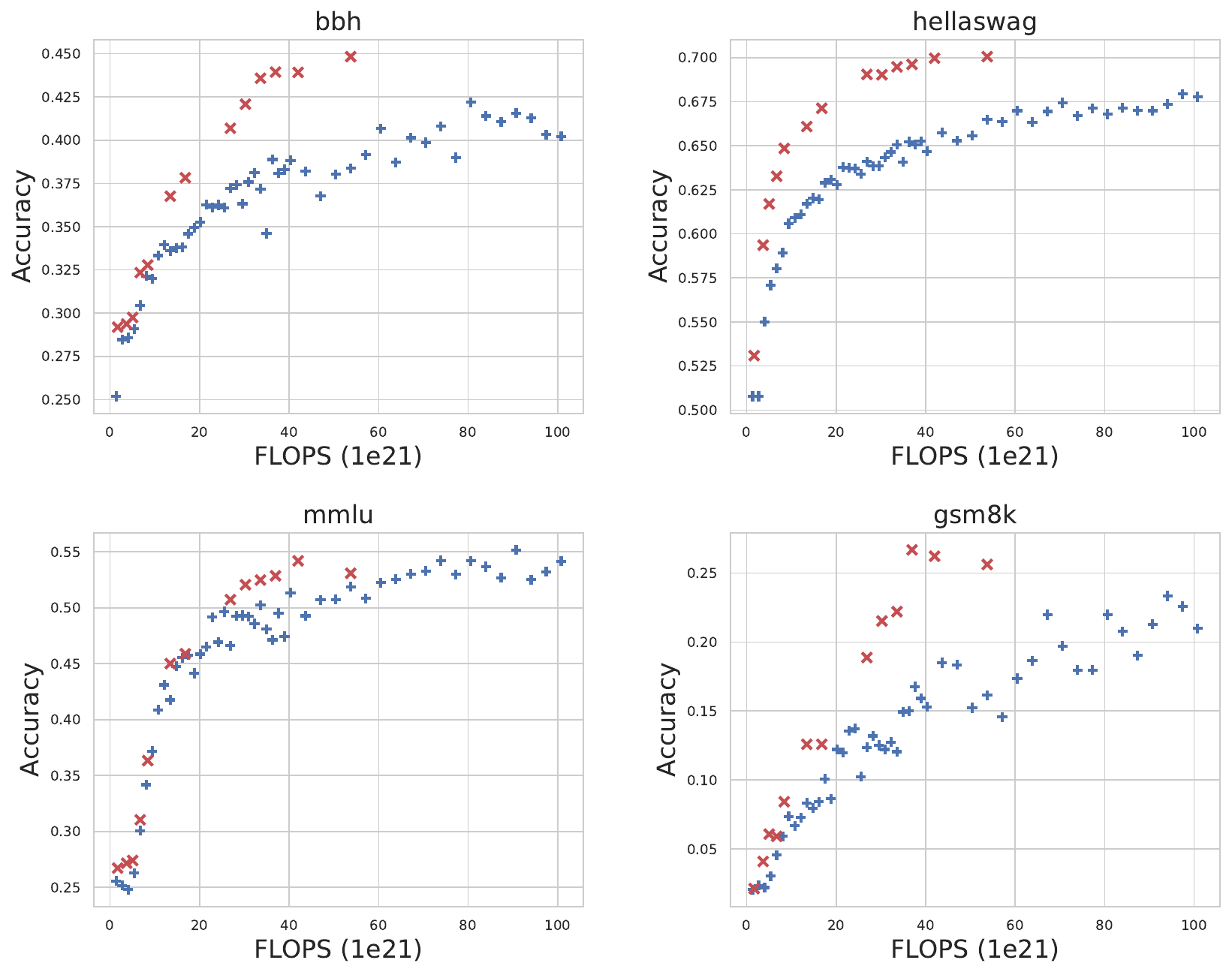}
\caption{Relationship between computation (FLOPs) and performance (accuracy) across various downstream tasks is depicted, with red and blue points indicating model checkpoints trained on two different data distributions. Under the same computational budget, models trained on different data distributions exhibit varying downstream task performance. }
\label{fig:computation-performance}
\end{figure}

Previous work on downstream scaling laws has primarily explored whether the performance of large models on downstream tasks can be predicted using pretrained small models. These studies follow a common setting, where small models trained on the same pretraining data distribution are used to predict the performance of larger models. However, during the pretraining process, researchers may need to continuously adjust the data composition to compare the impact of different data distributions on model performance. In this framework, applying downstream scaling laws would require training multiple small models under different configurations with the sa data distributions, making the experiments significantly more costly.

In this section, we first investigate whether downstream scaling laws can be directly modeled using computational cost. Specifically, we use InternLM2.5 as the base architecture and train two series of models on significantly different pretraining data distributions which is the same pretraining data sources as InternLM2.5. These sources include a diverse mix of web data, code data, math data, and other domain-specific datasets. We only adjusted the proportions of different subsets within this pretraining corpus. We then select checkpoints at different levels of computational cost to evaluate downstream performance.

Figure \ref{fig:computation-performance} illustrates the performance of two models with different training data distributions on various downstream tasks under different computational budgets. We observe that for tasks such as Hellaswag and BBH, models trained with the same computational cost exhibit differing downstream task performance. This indicates that a direct function from computation to downstream task performance cannot be reliably modeled. Therefore, when predicting downstream task performance, we must account for differences in the models' training data distributions. Moreover, language models exhibit emergent capabilities in certain tasks, making it difficult to predict their performance using smaller models trained on the same pretraining data distribution. Before reaching the emergence threshold—when computational resources are insufficient and loss remains relatively high—the performance of small models on downstream tasks becomes difficult to measure \cite{du2024understanding}.

This raises an important question: Is there a suitable metric that satisfies the following conditions?
\begin{itemize}
    \item The metric provides continuous measurements and is independent of emergent capabilities.
    \item The metric reflects differences in data distribution and can be predicted during pretraining.
\end{itemize}
Inspired by traditional scaling laws, we consider validation loss as a potential candidate. Validation loss is continuous and serves as an indicator of model capability. Therefore, we aim to model the relationship between validation loss and downstream performance. To explore this, we measure the downstream performance of different open-source models and their average token loss on task validation datasets. To avoid biases introduced by different tokenizers across models, we normalize the loss calculation at the character level.

\begin{figure}[t]
    \centering
    \subfigure{
        \begin{minipage}[t]{0.49 \linewidth}
            \centering
\includegraphics[width=1\linewidth]{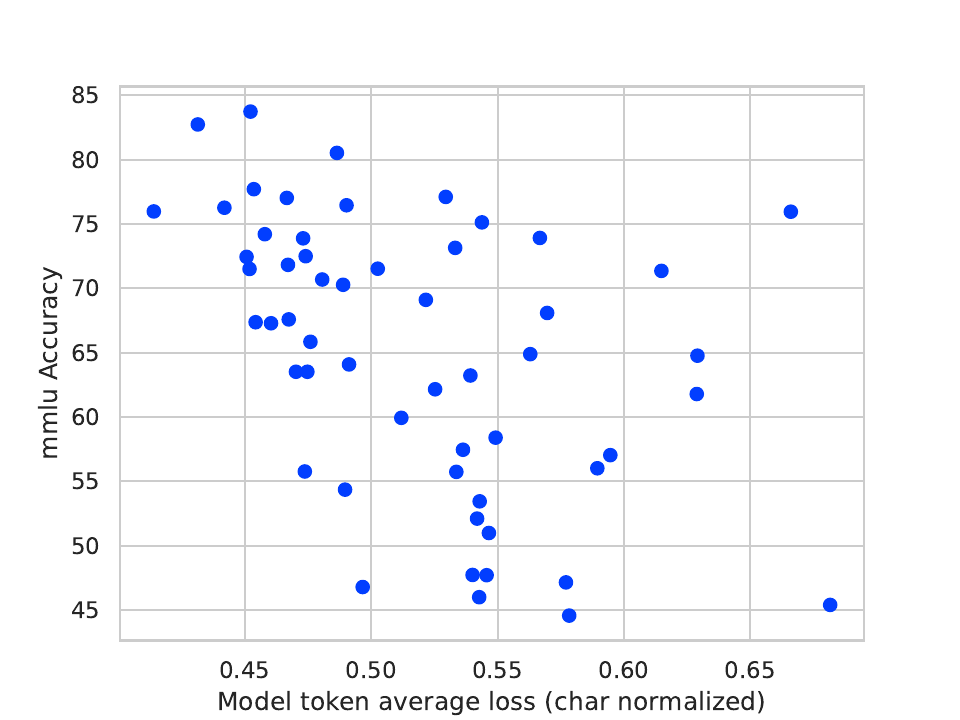}
        \end{minipage}
    }%
    \centering
    \subfigure{
        \begin{minipage}[t]{0.49 \linewidth}
            \centering
\includegraphics[width=1\linewidth]{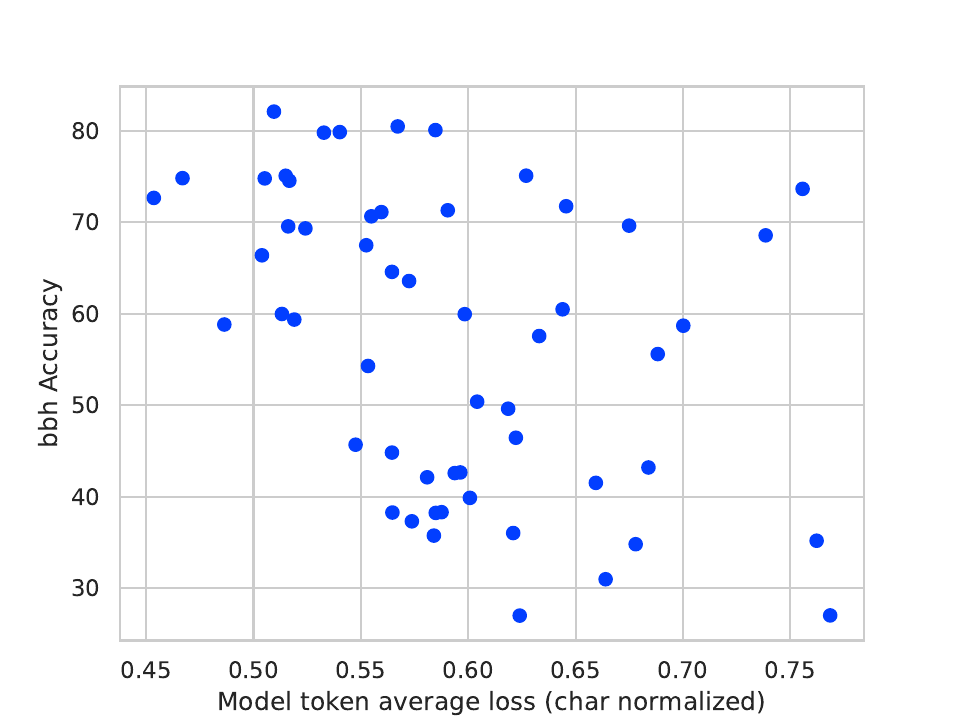}
        \end{minipage}
    }%

	\caption{The relationship between average token loss and downstream task performance.}
	\label{fig: tl}
\end{figure}
Figure \ref{fig: tl} illustrates the relationship between validation loss and downstream performance in MMLU and BBH. We observe that models can exhibit varying downstream performance across different tasks with the same validation loss. 
We revisit the computation of validation loss. Traditional scaling laws calculate validation loss by averaging it at the dataset level, treating all loss values with equal importance. However, different loss values may contribute unequally to measuring a model’s task specific capability. This indicates that a more fine-grained approach may be necessary for modeling loss effectively. In the next section, we introduce the concept of the \textbf{Capability Salience Vector} (CSV). This method assigns different importance weights to token-level loss values in the validation set, aiming to establish a more predictable relationship between validation loss and downstream performance.

\begin{figure*}[t]
\centering
\includegraphics[width=5.0in]{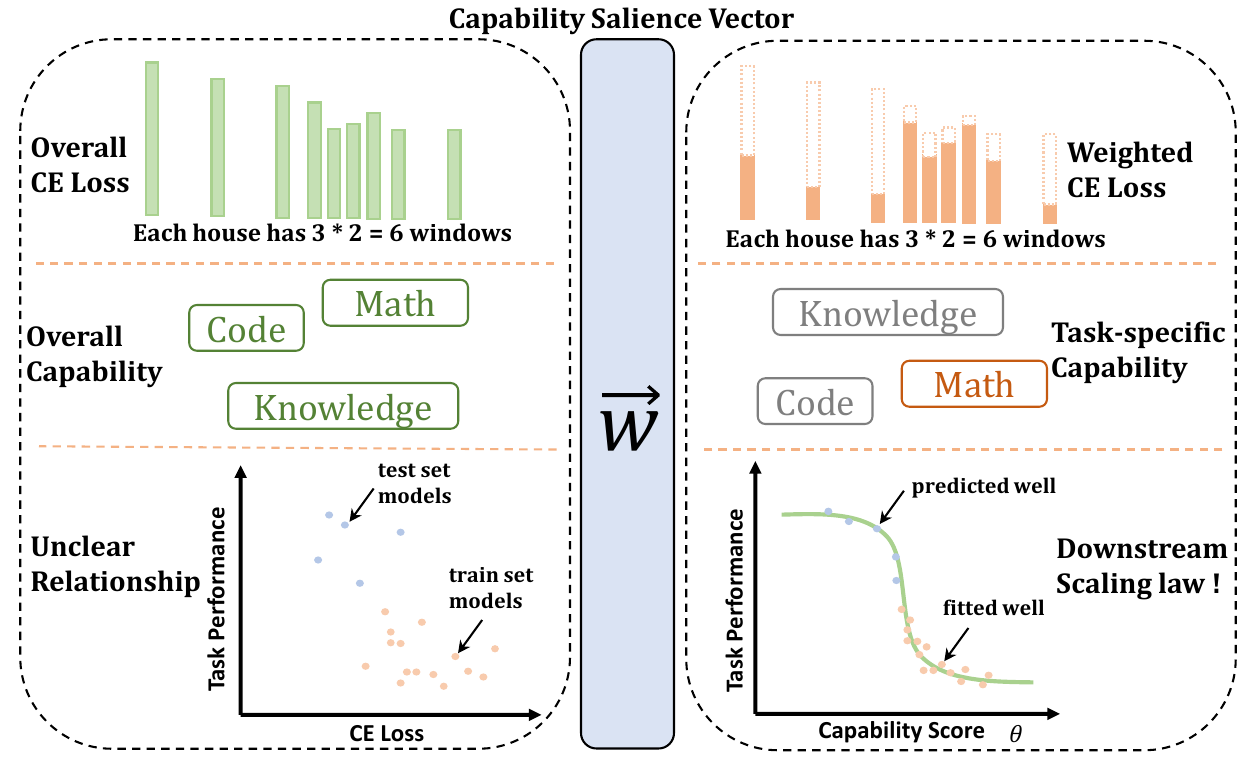}
\caption{Capability salience vector establishes the relationship between cross-entropy loss and downstream task performance. It quantifies a model's capabilities as a weighted sum of token losses, with each token contributing differently to the overall performance.
}
\label{fig:main}
\end{figure*}

\subsection{Modeling Downstream Scaling Law}

Inspired by the work of \citet{arora2023theory}., we consider that predicting different token fragments requires different meta-capabilities. These meta-capabilities ultimately determine the performance of downstream tasks. Traditional scaling laws treat all tokens equally when modeling validation loss. While the expectation of validation loss provides a rough estimate of model capability, it does not accurately capture the relationship between loss and downstream performance. To solve this problem, we propose the Capability Salience Vector. By applying a simple linear weighting to loss calculation, we can better model the relationship between loss and downstream performance. Given a validation set \( S = \{X_s\}_{s=1}^{|S|} \), we first compute the validation loss of each sample  \( X_s = (x_1, x_2, ..., x_{|X_s|}) \). Next, we use the \textbf{Capability Salience Vector}, denoted as \( W = \{w_{s,i}\} \), to apply linear weighting to each token's loss. This allows us to compute the capability score $\text{C}_{m}$ , which represents the model’s ability based on the validation text.

\begin{equation}
    \text{C}_{m} = - \frac{1}{N_{c}} \sum_{s=1}^{|S|} \sum_{i=1}^{|X_s|} w_{s,i} \log p(x_i \mid x_{<i}).
    \label{equ:score}
\end{equation}
where $N_{c}$ is the number of validation text characters. Then we use the sigmoidal function to model the functional relationship between capability score $\text{C}_{m}$ and the model's downstream task performance (e.g., accuracy) :
\begin{equation}
A_{t,m} = \gamma + \frac{1 - \gamma}{1 + \exp(-\alpha(\text{C}_{m} - \beta))}.
    \label{equ:dsl}
\end{equation}
where $\alpha$ and $\beta$ is the parameters to be fitted, $\gamma$ represents the expected performance of the task under random guessing.

\subsection{Modeling Capability Salience Vector} \label{sec3.3}
Since the weights of the Capability Salience Vector and the parameters $\alpha$ ,$\beta$ are unknown, we use three steps to get the weights of capability Salience vector.
\paragraph{Extract Weights of Capability Salience Vector}
For a given validation set, we first apply a loss mapping algorithm to align losses across different vocabulary spaces. Specifically, we average the token-level losses over their corresponding characters to obtain character-level losses, and then aggregate these character-level losses according to the tokenization result of the target tokenizer. Then we use language model with a scoring head to obtain  capability Salience Vector weights \( W = \{w_{s,i}\}\) of each token:
\begin{equation}
w_{s,i} = f_{\theta} (x_{s,i}|x_{s,<i} ).
\end{equation}
where $\theta$ is the parameters of scoring head. The scoring head is implemented as a linear layer applied to the final hidden states of a the language model and we froze the other parameters of language model during optimization. This step can be seen as assigning a score to each token, reflecting its contribution to the overall capability representation.
\paragraph{Fitting Downstream Scaling Law Function}
After getting capability Salience Vector, we collect token losses on the given validation set and downstream task performance of different models to fitting downstream scaling law function. To do this, we fix \( \theta \) and estimate the parameters \( \alpha \) and \( \beta \) of the downstream scaling law function. We minimize the MSE loss between the predicted and observed downstream task performance by using the Levenberg-Marquardt algorithm.
\begin{equation}
\min_{\alpha,\beta} \sum_{\text{Models } m} [ A_{t,m} - \hat{A}_{t,m}]^2.
\end{equation}

\paragraph{Optimize Capability Salience Vector}
In this stage, we use the downstream scaling law function to optimize Capability Salience Vector parameters $\theta$ . This is done by minimizing the MSE loss between predicted and observed downstream task performance using SGD: 
\begin{equation}
\min_{\theta} \sum_{\text{Models } m} [ A_{t,m} - \hat{A}_{t,m}]^2.
\end{equation}
We run multiple iterations of the three-steps optimization process to select the parameters that achieve the best downstream performance prediction. These optimal parameters are saved to obtain the \textbf{Capability Salience Vector} and fit downstream scaling law function. Finally, our algorithm workflow is illustrated in Appendix \ref{AP.A}.

\begin{figure*}[htbp]
\centering
\includegraphics[width=0.8\linewidth]{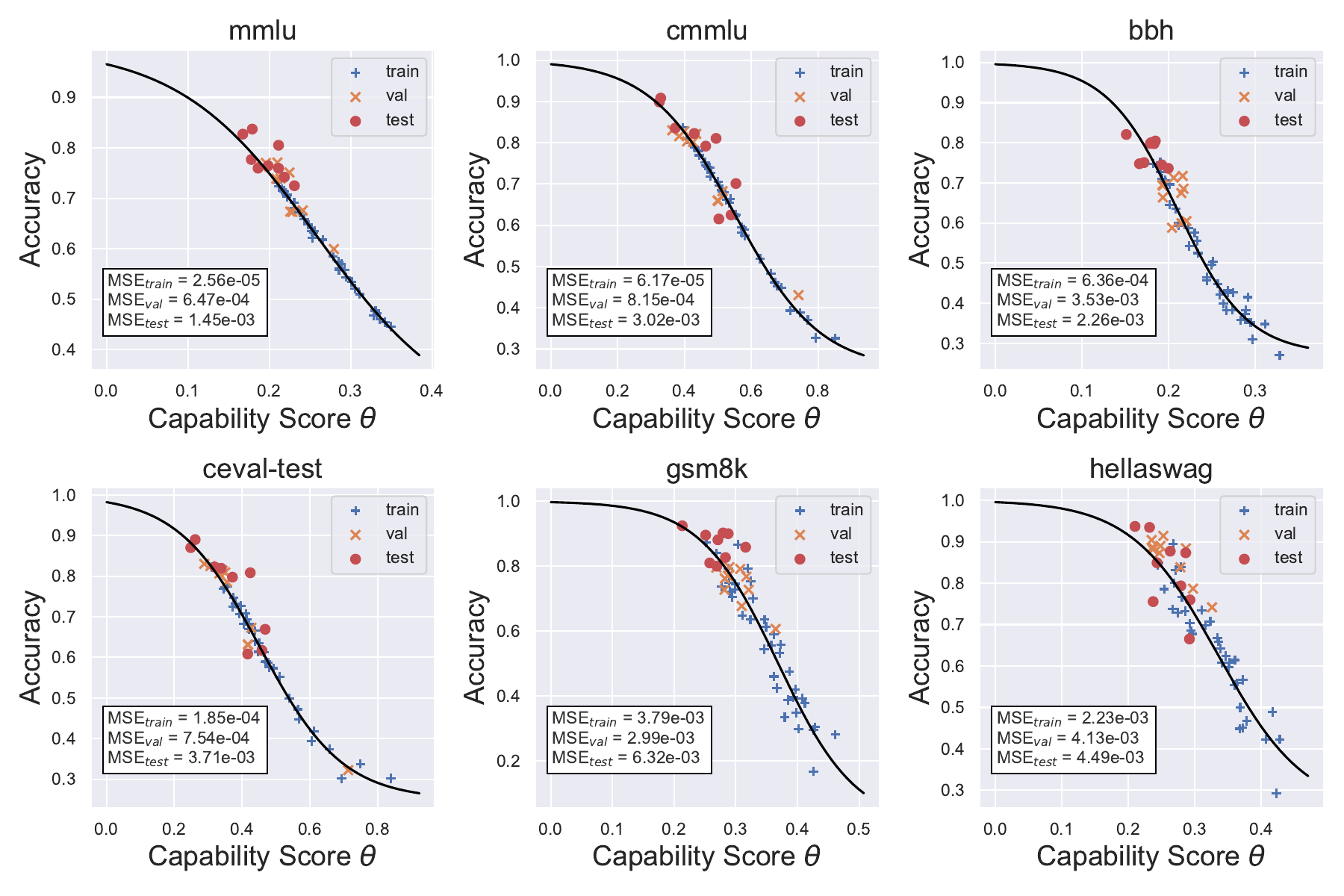}
\caption{Downstream task prediction results of open-source models using capability score derived from Capability Salience Vector.}
\label{fig:open_pred}
\end{figure*}

%% file: section/4_experiment.tex
\section{Experiment}
To validate the effectiveness of our approach, we divide the experiment into two parts. First, we evaluate the performance of the Capability Salience Vector using open-source models. Next, we use checkpoints from our own models to assess whether the Capability Salience Vector, optimized from open-source models, can accurately predict downstream task performance for models trained on different data distributions.

\subsection{Settings}
\paragraph{Dataset} To explore the scaling capabilities of models across various downstream tasks requiring different abilities, we conducted our experiments on six popular benchmarks that cover a wide range of skills, including knowledge, reasoning, commonsense, and mathematics. For knowledge capabilities, we choose MMLU \cite{hendrycks2020measuring}. For reasoning capabilities, we choose BBH \cite{suzgun2022challenging}. For commonsense capabilities, we choose Hellaswag \cite{zellers2019hellaswag}. For mathematics capabilities, we choose Gsm8k \cite{cobbe2021training}. For Chinese language capabilities, we selected the CMMLU \cite{li2023cmmlu} and Ceval \cite{huang2024c} benchmarks.

\paragraph{Evaluation Models} Our experiments were conducted under two different evaluation setups:

\begin{itemize}
    \item Open-source Model Evaluation: We validate the effectiveness of our method on over 50 open-source models across different series, including LLAMA2 \cite{touvron2023llama}, LLAMA3 \cite{dubey2024llama}, Gemma2 \cite{team2024gemma}, Qwen1.5 \cite{bai2023qwen}, Qwen2 \cite{yang2024qwen2}, Yi, Yi1.5\cite{young2024yi}, InternLM2, and InternLM2.5 \cite{cai2024internlm2}. In this setup, we use smaller models from the same series as the training set, while larger models are treated as the validation and test sets. The specific splits are detailed in Appendix \ref{Ap.C}.
    \item Closed-source Model Evaluation: We train two different series of models under varying data distributions and aim to predict their downstream capabilities using the Capability Salience Vector. The train data source is the same as InternLM2.5. For Capability Salience Vector optimization, we use all the open-source models along with early checkpoints (the first 50k steps) from both series. One series of our own model is used as the validation set, while the other serves as the test set.
\end{itemize}

\begin{figure*}[ht]
\centering
\includegraphics[width=0.8\linewidth]{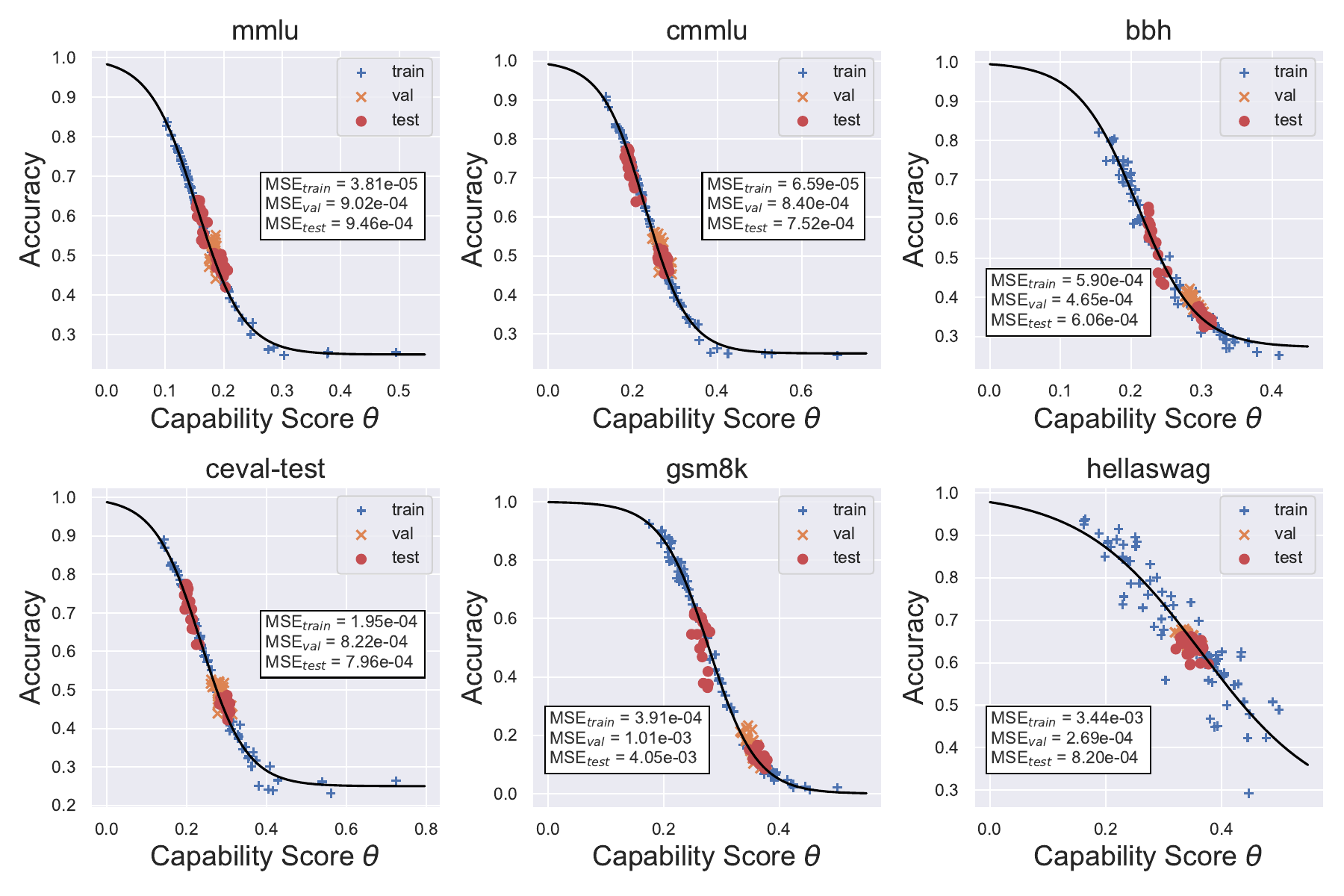}
\caption{Downstream task prediction results of closed-source models using capability score derived from Capability Salience Vector.}
\label{fig:closed_pred}
\end{figure*}

\paragraph{Evaluation Details} We use the OpenCompass \cite{2023opencompass} tool to evaluate the downstream scores for the aforementioned models. OpenCompass is a python package that supports various large language models and datasets for evaluation and benchmarking. The detailed evaluation results for the open-source models are provided in Appendix \ref{Ap.D}. To obtain the token cross-entropy loss on specific validation sets, we utilize LMDeploy\cite{2023lmdeploy} framework. To ensure the validation loss better reflects the different capabilities of the models, we randomly sample 100 examples from the six validation sets and mixed them. Additionally, we include 50 examples with chain-of-thought related to BBH. We treated CMMLU and CEval as reflecting the same capability and only sampled from one of these datasets. In total, we gathered 550 examples for optimizing the Capability Salience Vector.
Our method can use only 550 examples as validation set for modeling downstream scaling law.

\paragraph{Baseline} 
\begin{itemize}
\item All token loss : This method simply calculate the average of all token loss in validation set. Then We use the metric to model downstream scaling law by fitting a sigmoidal function.
\item Label token loss : Following the same setup as Llama3 \cite{dubey2024llama}, we compute the negative log-likelihood of the correct answers on each task's test set and calculate its expectation. We then use this metric to fit a sigmoidal function between fitting performance and downstream task performance.

\end{itemize}

\paragraph{Training Details} We use InternLM2.5 1.8B \cite{cai2024internlm2} as the scoring model, replacing its language head with a linear layer followed by a non-linear activation function. In practice, we found that increasing model complexity has little impact on the final results. To optimize efficiently, we freeze the language model parameters and only train the linear layer. We train with a learning rate of 1e-3, and most tasks converged to the best performance within 300 steps(training for 1-3 hours on a single A800.), demonstrating the practical efficiency of our method.

\begin{figure*}[ht]
\includegraphics[width=\linewidth]{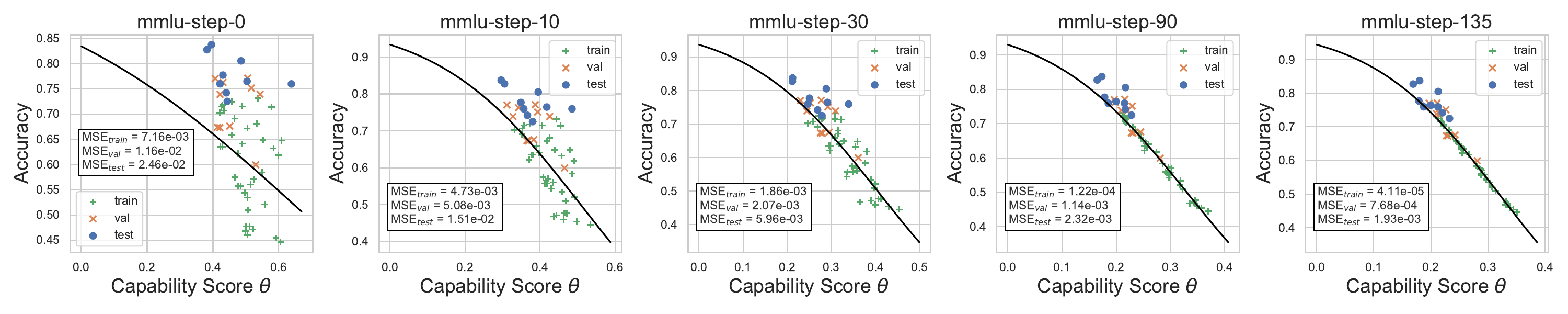}
\caption{Relationship between capability scores and downstream task performance across different steps of the Capability Salience Vector optimization process. As the optimization progresses, the capability scores of different model series gradually align, leading to improved predictive performance on the test set.}
\label{fig:dynamics}
\end{figure*}

\begin{table}[htbp]
\caption{MSE of Different loss to predict downstream performance.}\label{tab.loss}
\begin{tabular}{c|ccc|}
\hline
\hline
                       & \multicolumn{3}{c|}{Test Model Prediction MSE}                                                     \\ \cline{2-4} 
\multirow{-2}{*}{Task} & \multicolumn{1}{c|}{CSV}                            & \multicolumn{1}{c|}{All token} & Label token \\ \hline
mmlu                   & \multicolumn{1}{c|}{{\color[HTML]{FE0000} 1.45e-3}} & \multicolumn{1}{c|}{2.40e-2}   & 3.31e-2     \\
bbh                    & \multicolumn{1}{c|}{{\color[HTML]{FE0000} 2.26e-3}} & \multicolumn{1}{c|}{5.81e-2}   & 8.27e-2     \\
gsm8k                  & \multicolumn{1}{c|}{{\color[HTML]{FE0000} 6.32e-3}} & \multicolumn{1}{c|}{7.46e-2}   & 9.88e-2     \\
hellaswag & \multicolumn{1}{c|}{{\color[HTML]{FE0000} 4.49e-3}} & \multicolumn{1}{c|}{3.15e-3} & 3.33e-2 \\
ceval                  & \multicolumn{1}{c|}{{\color[HTML]{FE0000} 3.71e-3}} & \multicolumn{1}{c|}{2.67e-2}   & 3.57e-2     \\
cmmmlu    & \multicolumn{1}{c|}{{\color[HTML]{FE0000} 3.32e-3}} & \multicolumn{1}{c|}{2.69e-2} & 3.67e-2 \\ \hline\hline
\end{tabular}
\end{table}

\subsection{Main Results}

\begin{figure}[ht]
    \centering
    
    \hspace*{-0.2cm}
    \subfigure{
        \begin{minipage}[t]{0.33\linewidth}
\includegraphics[width=1\linewidth]{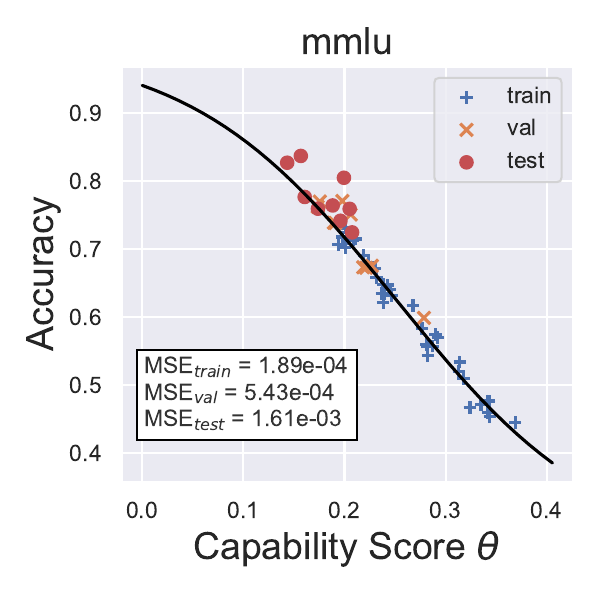}
        \end{minipage}
    }%
    \hspace*{-0.2cm}
    \subfigure{
        \begin{minipage}[t]{0.33\linewidth}
\includegraphics[width=1\linewidth]{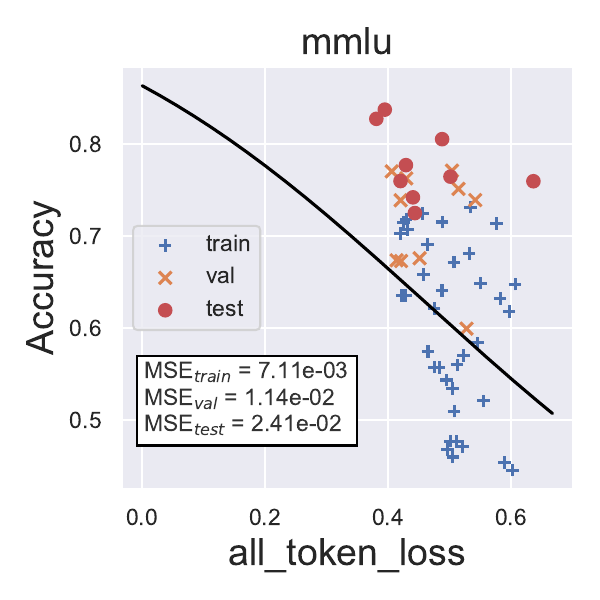}
        \end{minipage}
    }%
    \hspace*{-0.2cm}
    \subfigure{
        \begin{minipage}[t]{0.33\linewidth}
\includegraphics[width=1\linewidth]{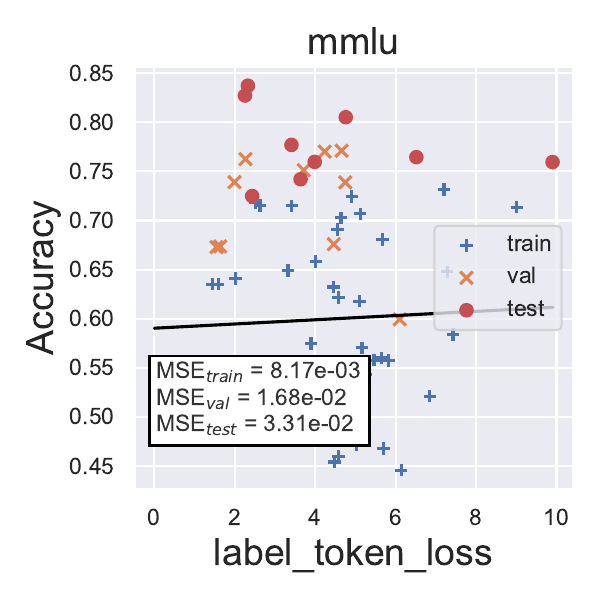}
        \end{minipage}
    }%
    \\
    
    \vspace{-0.5cm}
    
    \hspace*{-0.2cm}
    \subfigure{
        \begin{minipage}[t]{0.33\linewidth}
\includegraphics[width=1\linewidth]{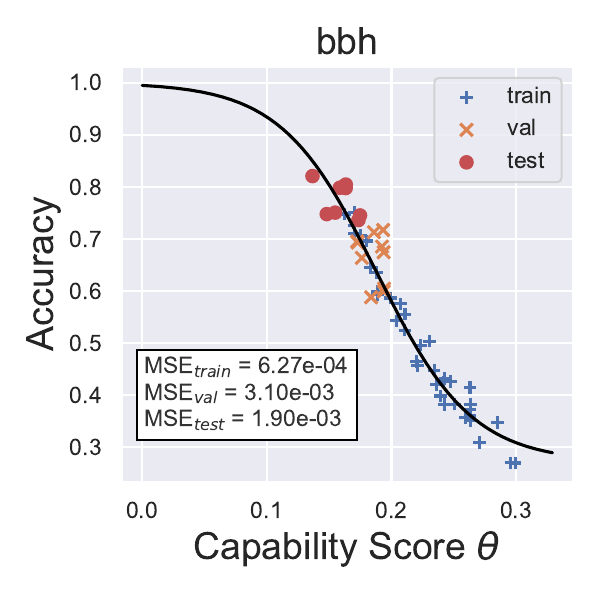}
        \end{minipage}
    }%
    \hspace*{-0.2cm}
    \subfigure{
        \begin{minipage}[t]{0.33\linewidth}
\includegraphics[width=1\linewidth]{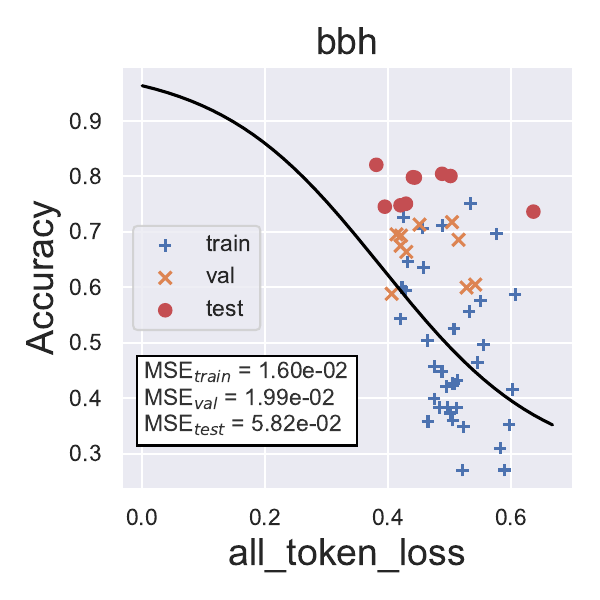}
        \end{minipage}
    }%
    \hspace*{-0.2cm}
    \subfigure{
        \begin{minipage}[t]{0.33\linewidth}
\includegraphics[width=1\linewidth]{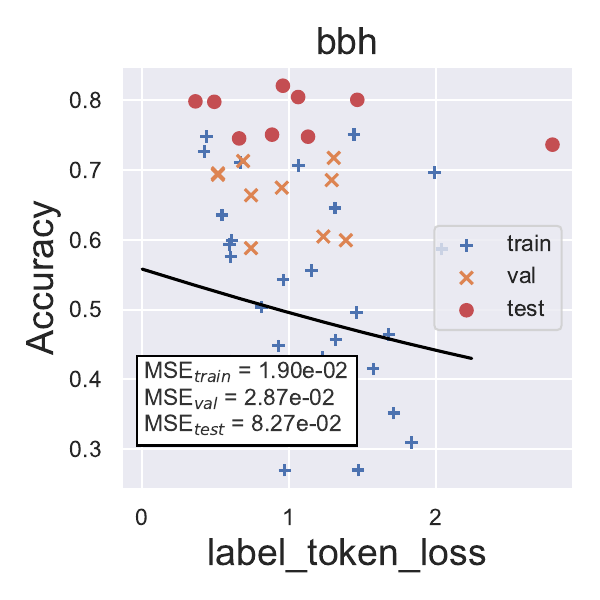}
        \end{minipage}
    }%

	\caption{Visualization of prediction results of our method compared to baseline}
	\label{fig: baseline}
\end{figure}

\paragraph{Downstream Task Prediction for Open-source Models} Figure \ref{fig:open_pred} shows the optimization results for different downstream tasks across various model families. It can be observed that by identifying a shared Capability Salience Vector across different model families on validation text, we are able to align the capabilities across different models for downstream tasks. Furthermore, the sigmoidal function fitted using the Capability Salience Vectors can effectively predict the downstream task performance of models with larger parameters. This indicates that the downstream task capabilities of models also follow a scaling law, which makes their performance predictable. Table \ref{tab.loss} and Figure \ref{fig: baseline} present the prediction MSE of different loss calculation methods. The capability salience vector significantly improves the predictability of downstream task performance by assigning different importance weights to different token losses. In contrast, the label token loss does not perform well. We believe this is because the loss at the position of the correct answers in the downstream task test set is susceptible to biases in the pretrain data distribution. As a result, this metric fails to align across different model families. We further validated this on checkpoints from our own trained models.

\paragraph{Downstream Task Prediction for Closed-source Models} The results in figure \ref{fig:closed_pred} demonstrate that by using the Capability Salience Vector as a proxy, we can predict downstream performance during the model training process, even when models are trained with different data distribution. During training, we were able to maintain a prediction accuracy for downstream task performance with a mean squared error of 1e-4. This demonstrates that our method provides strong guidance for model training and effectively predicts downstream performance.

\subsection{Convergence Trend of Capability Salience Vector Optimization}

Figure \ref{fig:dynamics} depicts the optimization trend of model capabilities during the process of Capability Salience Vector optimization. The figure shows that the model capabilities progressively converge toward a universal curve shared across different models. This convergence demonstrates that the LM cross-entropy loss reliably reflects model capabilities and that a functional relationship exists between LM cross-entropy loss and downstream task performance. Our algorithm successfully models the downstream scaling law across various model families, improving the predictability of downstream performance based on LM cross-entropy loss.

\subsection{Analysis Studies}

\begin{figure}[ht]
\includegraphics[width=\linewidth]{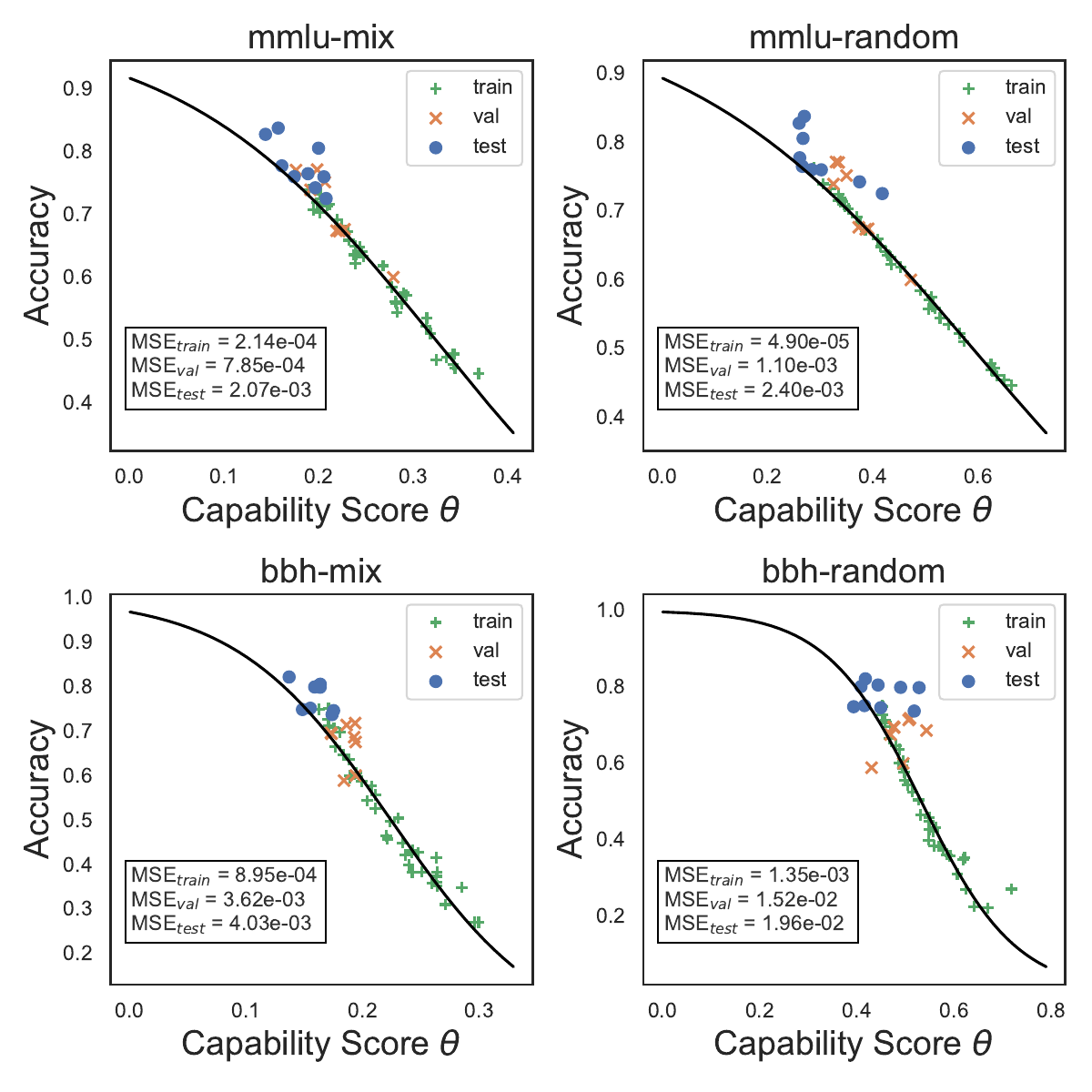}
\caption{Impact of varying validation set distributions on  Capability Salience Vector optimization.}
\label{fig:val-diff}
\end{figure}

\paragraph{Impact of Validation Set Data Distribution on Capability Salience Vector Optimization}  We found that as the diversity of the validation set distribution increases, the range of meta-capability combinations captured by the model from LM cross-entropy loss also expands. This leads to more accurate capability predictions for downstream scaling law. Therefore, when using loss to measure model capabilities, we recommend considering token losses from a more diverse set of data sources, rather than relying solely on the loss from a single data distribution. The result is shown in Figure \ref{fig:val-diff}

\begin{figure}[t]
\centering
\includegraphics[width=\linewidth]{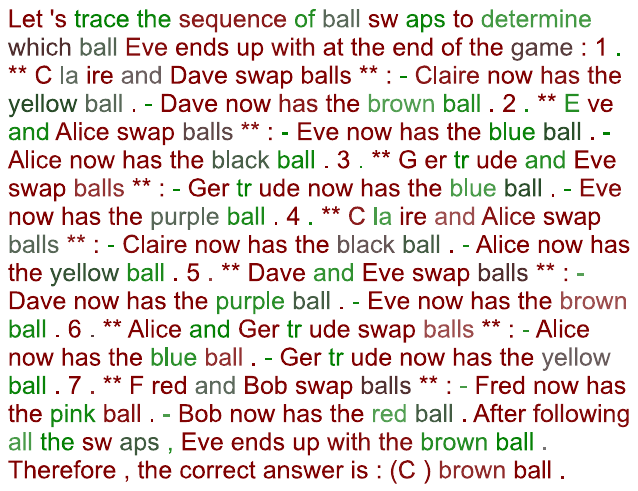}
\caption{Visualization of the importance weights assigned to different tokens by the Capability Salience Vector after optimization on the BBH task. Tokens with higher importance weights are highlighted in green. This shows that the Capability Salience Vector effectively identifies tokens that require reasoning by the model, capturing the model's reasoning capabilities.}
\label{fig:case}
\end{figure}

\paragraph{Case Study} In this section, we visualize the Capability Salience Vectors identified by the automatic optimization algorithm. Figure \ref{fig:case} shows the results of the Capability Salience Vector optimized for the BBH downstream task. The intensity of the green color represents the importance weight assigned by the Capability Salience Vector, with darker shades indicating larger weights. We observe that the optimized Capability Salience Vector highlights the tokens which can measure the model's reasoning capabilities. The model needs to summarize contextual information and apply reasoning to predict these tokens. These examples demonstrate that our method effectively extracts meta-capabilities that capture the model's scaling behavior on specific tasks. 

%% file: section/5_conclusion.tex
\section{Conclusion}
In this work, we first demonstrated that different tokens correspond to distinct meta-capabilities, highlighting the importance of accounting for token-level differences when exploring the scaling relationship between LM CE loss and downstream task performance. Furthermore, we proposed an automatic optimization method to identify the Capability Salience Vector for a text segment. By assigning varying importance weights to the losses of different tokens, our approach enables researchers to model the scaling relationship between loss and downstream tasks. We believe that precisely modeling the scaling relationship between language CE loss and downstream task performance is a meaningful and valuable area for future research.

%% file: section/6_limitation.tex
\section{Limitation}
In this section, we will discuss some limitations of our method. First, while our approach can optimize any given text to identify its Capability Salience Vector, the effectiveness varies across different texts. Therefore, selecting suitable optimization texts remains an open area for future research. Second, since the optimization needs to be performed simultaneously on all texts, the computation cost increases with the length of the text. Thus, improving the efficiency of the optimization algorithm or exploring simpler modeling approaches is a topic for future investigation. Finally, our experiments are limited to six popular objective evaluation benchmarks. Exploration of more challenging tasks or subjective evaluations is left for future work.

\section*{Acknowledgements}
The authors wish to thank the AC and anonymous reviewers for their constructive comments. This work was supported by the China Postdoctoral Science Foundation under Grant Number 2023M741851.

%% file: section/appendix.tex
\newpage
\appendix

\section{Appendix}
\addcontentsline{toc}{section}{Appendix}
\renewcommand{\thesubsection}{\Alph{subsection}}

\subsection{Open-source Model Evaluation Setting} \label{Ap.C}

In this section, we present the specific model splits used for Capability Salience Vector optimization in the evaluation of open-source models. Larger models within the same series were used for validation and testing, as detailed in Table \ref{tab.2}. This split ensures that our method is evaluated across models of varying sizes while maintaining consistency within each series for training and validation.

\subsection{Algorithm Workflow}\label{AP.A}

Based on the derivation in Section \ref{sec3.3}, we summarise the algorithmic flow of Capability Salience Vector modeling in Algorithm \ref{alg:loss_mapping} and \ref{alg:Capability_Vector_Modeling}.

\begin{table*}[ht]
\caption{Model split in open-source model evaluation}\label{tab.2}
\centering
\begin{tabular}{lll}
\hline\hline
\textbf{train}                   & \textbf{validation}      & \textbf{test}         \\ \hline
Llama-2-\{7, 13\}B               & Llama-2-70B              & Yi-1.5-34B            \\
Llama-2-\{7, 13\}B-chat          & Llama-2-70B-chat         & Llama-3-70B           \\
Qwen1.5\{1.8, 4, 7, 14\}B        & Qwen1.5-\{32, 72\}B      & Llama-3-70B-Instruct  \\
Qwen1.5\{1.8, 4, 7, 14\}B-chat   & Qwen1.5-\{32, 72\}B-Chat & Yi-1.5-34B-Chat       \\
Llama3-8B                        & internlm2-chat-20B       & gemma-2-27B           \\
Llama3-8B-Instruct               & internlm2-20B            & gemma-2-27B-it        \\
Qwen2\{1.5, 7\}B                 & internlm2-chat-20B-sft   & Qwen2-72B             \\
Qwen2\{1.5, 7\}B-Instruct        & Yi-\{34\}B                & Qwen2-72B-Instruct    \\
Yi-\{6\}B                        & Yi-\{34\}B-chat           & internlm2\_5-20B      \\
Yi-\{6\}B-chat                   &                          & internlm2\_5-20B-chat \\
Yi1.5-\{6, 9\}B                  &                          & Yi-1.5-34B-Chat       \\
Yi1.5-\{6, 9\}B-chat             &                          &                       \\
Gemma2-2-\{2, 9\}B               &                          &                       \\
internlm2-\{1.8, 7\}B            &                          &                       \\
internlm2-\{1.8, 7\}B-chat       &                          &                       \\
internlm2-\{1.8, 7\}B-chat-sft   &                          &                       \\
internlm2.5-\{1.8, 7\}B          &                          &                       \\
internlm2.5-\{1.8, 7\}B-chat     &                          &                       \\
internlm2.5-\{1.8, 7\}B-chat-sft &                          &                       \\ \hline\hline
\end{tabular}
\end{table*}

\begin{algorithm*}
\caption{Loss Mapping Between Different Tokenizers}
\label{alg:loss_mapping}
\begin{algorithmic}[1]

\REQUIRE Input text $C = (c_1, c_2, \dots, c_n)$
\REQUIRE Source tokenizer $T_s$ and target tokenizer $T_t$
\REQUIRE Source token losses $L_s = (\ell_1^s, \ell_2^s, \dots, \ell_m^s)$

\STATE Tokenize input with source tokenizer: $T_s(C) = (\tau_1^s, \dots, \tau_m^s)$
\STATE Initialize character-level losses $\ell_j^c = 0$ for each character $c_j \in C$

\FOR{each token $\tau_i^s$ in $T_s(C)$}
    \STATE Let $S_i$ be the indices of characters spanned by $\tau_i^s$
    \FOR{each $j \in S_i$}
        \STATE $\ell_j^c =  \frac{\ell_i^s}{|\tau_i^s|}$
    \ENDFOR
\ENDFOR

\STATE Tokenize input with target tokenizer: $T_t(C) = (\tau_1^t, \dots, \tau_k^t)$
\STATE Initialize target token losses $L_t = (\ell_1^t, \dots, \ell_k^t)$

\FOR{each token $\tau_j^t$ in $T_t(C)$}
    \STATE Let $T_j$ be the indices of characters spanned by $\tau_j^t$
    \STATE $\ell_j^t = \sum_{i \in T_j} \ell_i^c$
\ENDFOR

\RETURN Target token losses $L_t$
\end{algorithmic}
\end{algorithm*}

\begin{algorithm*}
\caption{Optimization of Capability Salience Vector and Downstream Scaling Law}
\label{alg:capability_optimization}
\begin{algorithmic}[1]
\REQUIRE Validation set $S = \{X_s\}_{s=1}^{|S|}$
\REQUIRE Language model with score head $f_{\theta}$
\REQUIRE Model Set $M = \{m\}$
\REQUIRE Observed downstream performance of different models $\hat{A}_{t,m}$

    \STATE Collect token losses on the validation set of different model $\{L_m\}$ and mapping the losses from source tokenizer to target tokenizer Using algorithm \ref{alg:loss_mapping}
    
    \FOR{each epoch}

        \STATE \textbf{Step 1: Extract Weights of Capability Salience Vector}
        \FOR{each sample $X_s = (x_1, x_2, ..., x_{|X_s|}) \in S$}
            \FOR{each token $x_{s,i}$ in $X_s$}
                \STATE Compute $w_{s,i} = f_{\theta}(x_{s,i} \mid x_{s,<i})$ using the language model scoring head.
            \ENDFOR
        \ENDFOR

        \STATE Obtain initial Capability Salience Vector $W = \{w_{s,i}\}$.
        
        \FOR{each model $m$ in Model Set $M$}
            \STATE \textbf{Step 2: Fit Downstream Scaling Law Function}
            \STATE Using equation \ref{equ:score} and \ref{equ:dsl} to predict downstream task performance $A_{t,m}$.
            \STATE Fix $\theta$ and estimate $\alpha, \beta$ by minimizing the MSE loss:
            \begin{equation*}
            \min_{\alpha,\beta} \sum_{m} [ A_{t,m} - \hat{A}_{t,m}]^2
            \end{equation*}
            \STATE Use the Levenberg-Marquardt algorithm to solve for $\alpha, \beta$.
            
            \STATE \textbf{Step 3: Optimize Capability Salience Vector}
            \STATE Using equation \ref{equ:score} and \ref{equ:dsl} to predict downstream task performance $A_{t,m}$.
            \STATE Update $\theta$ by minimizing:
            \begin{equation*}
            \min_{\theta} \sum_{m} [ A_{t,m} - \hat{A}_{t,m}]^2
            \end{equation*}
            \STATE Use Stochastic Gradient Descent (SGD) to optimize $\theta$.
        \ENDFOR
        
        \STATE Repeat Steps 1-3 until convergence.
    \ENDFOR
    
\RETURN Optimized Capability Salience Vector $W$ and Downstream Scaling Law parameters $\alpha, \beta$.
\end{algorithmic}
\label{alg:Capability_Vector_Modeling}
\end{algorithm*}

\subsection{Open-source Model Evaluation Results} \label{Ap.D}
In this section, we present the evaluation results in Table \ref{tab.3} for various tasks on open-source models using OpenCompass.

\begin{table*}[ht]
\caption{Downstream Tasks Evaluation Results(1)}\label{tab.3}
\centering
\begin{tabular}{ccccccc}
\hline\hline
model                     & bbh   & ceval-test & cmmlu & gsm8k & hellaswag & mmlu  \\ \hline
Llama-2-7b-hf             & 38.27 & 30.13      & 32.75 & 16.76 & 29.29     & 46.78 \\
Llama-2-13b-hf            & 45.68 & 37.38      & 38.81 & 29.87 & 45.06     & 55.76 \\
llama-2-70b-hf            & 64.78 & 49.53      & 53.05 & 63.53 & 55.91     & 69.87 \\
Qwen1.5-1.8B              & 27.01 & 58.67      & 57.45 & 34.87 & 42.32     & 47.14 \\
Qwen1.5-4B                & 34.81 & 66.55      & 66.38 & 47.61 & 55.89     & 57.03 \\
Qwen1.5-7B                & 39.87 & 72.49      & 71.86 & 54.36 & 68.51     & 62.15 \\
Qwen1.5-14B               & 50.38 & 76.93      & 76.95 & 63.53 & 83.86     & 69.1  \\
Qwen1.5-32B               & 67.47 & 82.5       & 81.58 & 72.71 & 87.28     & 73.88 \\
Qwen1.5-72B               & 58.81 & 83.03      & 83    & 79.53 & 90.41     & 77.02 \\
Qwen2-7B                  & 54.28 & 82.22      & 83.67 & 73.77 & 73.74     & 70.27 \\
Qwen2-72B                 & 74.52 & 89         & 90.82 & 89.54 & 93.46     & 83.73 \\
Yi-6B                     & 44.82 & 70.78      & 74.01 & 39.58 & 66.83     & 64.08 \\
Yi-34B                    & 66.37 & 80.93      & 82.79 & 67.7  & 83.83     & 76.26 \\
Yi-1.5-6B                 & 57.55 & 66.93      & 69.68 & 61.33 & 70.79     & 64.88 \\
Yi-1.5-9B                 & 71.09 & 72.71      & 74.11 & 74.53 & 76.64     & 71.52 \\
Yi-1.5-34B                & 75.06 & 82.24      & 83.55 & 79.98 & 84.92     & 77.7  \\
Qwen1.5-1.8B-Chat         & 27.03 & 55.19      & 48.3  & 29.57 & 42.32     & 45.39 \\
Qwen1.5-4B-Chat           & 43.19 & 61.37      & 58.22 & 46.02 & 60.74     & 56.01 \\
Qwen1.5-7B-Chat           & 35.19 & 68.18      & 67.98 & 55.88 & 69.8      & 61.78 \\
Qwen1.5-14B-Chat          & 55.58 & 74.67      & 75.29 & 64.82 & 80.03     & 68.08 \\
Qwen1.5-32B-Chat          & 68.55 & 80.66      & 80.17 & 79.15 & 88.42     & 75.12 \\
Qwen1.5-72B-Chat          & 71.73 & 81.37      & 82.15 & 79.68 & 88.99     & 77.1  \\
Qwen2-7B-Instruct         & 64.56 & 82.11      & 80.84 & 87.29 & 78.57     & 70.68 \\
Qwen2-72B-Instruct        & 82.07 & 87.02      & 89.88 & 92.42 & 93.7      & 82.73 \\
Yi-6B-Chat                & 30.98 & 70.69      & 72.74 & 42.46 & 64.27     & 63.22 \\
Yi-34B-Chat               & 60.47 & 78.58      & 80.33 & 72.71 & 78.69     & 73.91 \\
Yi-1.5-6B-Chat            & 58.68 & 68.05      & 68.59 & 75.36 & 75.66     & 64.76 \\
Yi-1.5-9B-Chat            & 69.61 & 73.19      & 74.51 & 79.23 & 83.14     & 71.35 \\
Yi-1.5-34B-Chat           & 73.63 & 80.83      & 81.05 & 85.82 & 87.33     & 75.95 \\
internlm2-1\_8b           & 36.03 & 44.79      & 45.27 & 30.4  & 44.86     & 45.99 \\
internlm2-7b              & 63.56 & 63.54      & 66.17 & 69.98 & 89.52     & 65.84 \\
internlm2-20b             & 71.29 & 67.28      & 68.28 & 76.8  & 91.41     & 67.58 \\
internlm2-chat-1\_8b      & 37.31 & 47.06      & 46.19 & 39.2  & 59.71     & 47.72 \\
internlm2-chat-1\_8b-sft  & 38.23 & 47.21      & 46.17 & 37.83 & 60.86     & 47.7  \\
internlm2-chat-7b         & 59.96 & 58.85      & 62.66 & 72.1  & 84.44     & 63.51 \\
internlm2-chat-7b-sft     & 59.35 & 58.99      & 62.57 & 70.51 & 85.01     & 63.51 \\
internlm2-chat-20b        & 69.53 & 63.03      & 65.85 & 76.19 & 88.16     & 67.36 \\
internlm2-chat-20b-sft    & 69.31 & 63.2       & 66.04 & 77.03 & 88.68     & 67.28 \\
internlm2\_5-1\_8b-chat   & 42.58 & 61.25      & 62.3  & 53.3  & 49.97     & 50.98 \\
internlm2\_5-7b           & 72.64 & 77.21      & 78.95 & 74.83 & 73.49     & 71.5  \\
internlm2\_5-7b-chat      & 74.8  & 77.38      & 78.04 & 84    & 67.74     & 71.82 \\
internlm2\_5-20b-chat     & 79.77 & 79.75      & 79.18 & 88.1  & 66.53     & 72.49 \\
internlm2\_5-1\_8b        & 42.65 & 63.97      & 65.5  & 38.82 & 56.64     & 53.44 \\
internlm2\_5-20b          & 79.83 & 81.94      & 82.25 & 82.64 & 79.35     & 74.2  \\ \hline\hline
\end{tabular}
\end{table*}

\begin{table*}[t]
\caption{Downstream Tasks Evaluation Results(2)}\label{tab.4}
\centering
\begin{tabular}{ccccccc}

\hline\hline
model                     & bbh   & ceval-test & cmmlu & gsm8k & hellaswag & mmlu  \\ \hline
Meta-Llama-3-8B           & 59.69 & 48.83      & 50.95 & 54.28 & 50.86     & 66.43 \\
Meta-Llama-3-70B          & 79.16 & 66.56      & 68.36 & 69.98 & 80.6      & 79.35 \\
Llama-2-7b-chat-hf        & 41.51 & 30.2       & 32.62 & 28.13 & 48.94     & 44.56 \\
Llama-2-13b-chat-hf       & 49.61 & 33.76      & 37.01 & 42.08 & 61.43     & 52.1  \\
Meta-Llama-3-8B-Instruct  & 52.5  & 49.93      & 51.89 & 79.3  & 73.29     & 67.16 \\
Meta-Llama-3-70B-Instruct & 80.45 & 66.91      & 70.11 & 90.22 & 87.72     & 80.52 \\
gemma-2-2b-it             & 46.44 & 41.82      & 44.86 & 55.65 & 62.43     & 58.39 \\
gemma-2-9b-it             & 75.07 & 57.32      & 59.49 & 86.66 & 70.29     & 73.14 \\
gemma-2-27b-it            & 80.05 & 61.71      & 62.5  & 89.99 & 75.96     & 76.45 \\
gemma-2-27b               & 74.77 & 60.84      & 61.53 & 80.97 & 75.55     & 75.97 \\
gemma-2-2b                & 42.12 & 39.44      & 39.31 & 33.51 & 65.91     & 54.35 \\
gemma-2-9b                & 70.63 & 57.63      & 59.02 & 72.86 & 72.93     & 72.44 \\
Llama-2-70b-chat-hf       & 59.94 & 32.21      & 43.07 & 60.65 & 74.18     & 59.93 \\
Qwen2-1.5B-Instruct       & 38.31 & 68.27      & 68.12 & 63.53 & 55.38     & 55.73 \\
Qwen2-1.5B                & 35.75 & 69.32      & 70.58 & 59.06 & 46.76     & 57.45 \\ \hline\hline
\end{tabular}
\end{table*}